\documentclass{article}

\usepackage[final, nonatbib]{neurips_2019}

\usepackage[utf8]{inputenc} 
\usepackage[T1]{fontenc}    
\usepackage{hyperref}       
\usepackage{url}            
\usepackage{booktabs}       
\usepackage{amsfonts}       
\usepackage{nicefrac}       
\usepackage{microtype}      
\usepackage[pdftex]{graphicx} 
\usepackage[ruled]{algorithm2e}
\usepackage{amsmath}
\usepackage{ulem}
\usepackage{float}

\usepackage[dvipsnames]{xcolor}

\title{FedMD: Heterogenous Federated Learning \\  via Model Distillation}
\author{%
Daliang Li \textsuperscript{1}, \;
Junpu Wang \textsuperscript{2, 3} \\
\textsuperscript{1}{Center for Fundamental Laws of Nature, 
  Harvard University}\\
\textsuperscript{2}{High Energy Theory Group, Yale University}\\
\textsuperscript{3}{Center for Particle Cosmology, University of Pennsylvania}\\
\texttt{daliang\_li@fas.harvard.edu},  \; \texttt{junpuwang@gmail.com}
}

\begin{document}

\maketitle

\begin{abstract}
Federated learning enables the creation of a powerful centralized model without compromising the data privacy of multiple participants.
While successful, it does not incorporate the case where each participant independently designs its own model. 
Due to intellectual property concerns and heterogeneous nature of tasks and data, this is a widespread requirement in applications of federated learning to areas such as health care and AI as a service. 
In this work, we use transfer learning and knowledge distillation to develop a universal framework that enables federated learning when each agent owns not only their private data, but also uniquely designed models. 
We test our framework on the MNIST/FEMNIST dataset and the CIFAR10/CIFAR100 dataset and observe fast improvement across all participating models. With 10 distinct participants, the final test accuracy of each model on average receives a $20\%$ gain on top of what's possible without collaboration and is only a few percent lower than the performance each model would have obtained if all private datasets were pooled and made directly available for all participants. 

\end{abstract}

\section{Introduction}
Deep learning has provided a potentially powerful framework to automate perception and inference. 
However, large datasets are required to fully realize this potential. 
In areas like health care, it is often difficult and costly to curate large datasets.
For instance, typical hospitals in the US may have only dozens of MRI images of a particular disease that needs to be annotated by human experts and must be protected from potential privacy breaches. 
Federated learning and similar ideas \cite{mcmahan2016FedAvg, Shokri:2015} rise to this challenge and effectively train a centralized model while keeping users' sensitive data on device. 
In particular, federated learning~\cite{mcmahan2016FedAvg, bonawitz2019towards,fedai} is optimized for faster communication and is uniquely capable of handling a large number of users. 

Federated learning faces many challenges~\cite{1908.07873}, among which, of particular importance is the heterogeneity that appear in all aspects of the learning process.
There is system heterogeneity when each participant has a different amount of bandwidth and computational power; 
this was partly resolved by the native asynchronous scheme of federated learning, which was further refined e.g. to enable active sampling~\cite{nishio2018client,kang2019incentive} and improve fault tolerance~\cite{sahu2018convergence}. 
There is also statistical heterogneity (the non i.i.d. problem) where clients have a varying amount of data coming from distinct distributions~\cite{chen2018federated, fed_multitask_smith_2017,varia_mtl,khodak2019adaptive,eichner2019semi,zhao2018federated}.

In this work, we focus on a different type of heterogeneity: the differences of local models. 
In the original federated framework, all users have to agree on the particular architecture of a centralized model. 
This is a reasonable assumption when the participants are millions of low capacity devices such as cell phones.
In this work, we instead explore extensions to the federated framework that is realistic in a business facing setting, where each participant has capacity and desire to design their own unique model. 
This arise in areas like health care, finance, supply chain and AI services. 
For example, when several medical institutions collaborate without sharing private data, they may need to craft their own model to meet distinct specifications. 
They may not be willing to share details of their models due to privacy and intellectual property concerns. 
Another example is AI as a service. A typical AI vendor of,  e.g. customer service chat bots,  may have dozens of client companies. 
Each client's model is distinct and solves different tasks.
The standard practice is to train a client's model with only its own data.
It would be immensely beneficial if data from other clients can be utilized without compromising privacy or independency. 
How can one perform federated learning when each participant has a different model that is a blackbox to others? 
This is the central question that we will answer in this work. 

This question is intimately related to the non-i.i.d. challenge of federated learning because a natural way to tackle statistical heterogeneity is to have individualized models for each user. Indeed, existing frameworks result in sightly different models. For example, \cite{fed_multitask_smith_2017} provides a framework for multi-task learning if the problem is convex. Approaches based on frameworks such as Bayesian \cite{varia_mtl}, meta-learning \cite{khodak2019adaptive} and transfer learning \cite{zhao2018federated} also achieve good performance on non-i.i.d. data while allowing a certain amount of model customization. However, to our knowledge, all existing frameworks require a centralized control over the design of local models. Full model independency, while related to the non-i.i.d. problem, is an important new research direction in its own right. 

The key to full model heterogeneity is communication. In particular, there must be a translation protocol enabling a deep network to understand the knowledge of others without sharing data or model architecture. 
This question touches on fundamental issues in deep learning, such as interpretability and emergent communication protocols. 
In principle, machines should be able to learn the best communication protocol that is adaptive to any specific use case. 
As a first step in this direction, we employ a more transparent framework based on knowledge distillation that solves the problem. 

Transfer learning is another major framework addressing the scarcity of private data. 
In this work, our private datasets can be as small as a few samples per class. 
Therefore using transfer learning from a large public dataset is imperative in addition to federated learning. 
We leverage the power of transfer learning in two ways. 
First, before entering the collaboration, each model is fully trained first on the public data and then on its own private data. 
Second, and more importantly,  the blackbox models communicate based on their output class scores on samples from the public dataset. 
This is realized through knowledge distillation~\cite{Hinton44873}, which has been capable of transmitting learned information in a model agnostic way. 
\vspace*{-0.1in}
\paragraph*{Contributions: }
The primary contribution of this work is FedMD, a new federated learning framework that enables participants to independently design their models. 
Our centralized server does not control the architecture of these models and only requires limited black box access.
We identify the key element of this framework to be the communication module that translates knowledge between participants.
We implement such a communication protocol by leveraging the power of transfer learning and knowledge distillation. 
We test this framework using a subset of the FEMNIST dataset~\cite{caldas2018leaf} and the CIFAR10/CIFAR100 datasets \cite{CIFAR}. We find significant gains in performance of local models using this framework compared to what's possible without collaboration. 

\section{Methods}
We propose the following challenge:
\subsection{Problem definition}
\label{sec:ProblemDefinition}
There are $m$ participants in the federated learning process. Each owns a very small labeled dataset $\mathcal{D}_k := \{(x_{i}^k, y_{i})\}_{i=1}^{N_k}$ that may or may not be drawn from the same distribution.
There is also a large public dataset $\mathcal{D}_0 := \{(x_{i}^0, y_{i}^0)\}_{i=1}^{N_0}$ that everyone can access.  
Each participant independently designs its own model $f_k$ to perform a classification task. 
The models $f_k$ can have different architectures. Furthermore, hyper-parameters need not to be shared among participants.
The goal is to establish a framework of collaboration that improves the performance of $f_k$ beyond individual effort with locally accessible data $\mathcal{D}_0$ and $\mathcal{D}_k$.

\vspace{-20pt}
\begin{figure}
  \centering
  
  \includegraphics[width=0.5\linewidth]{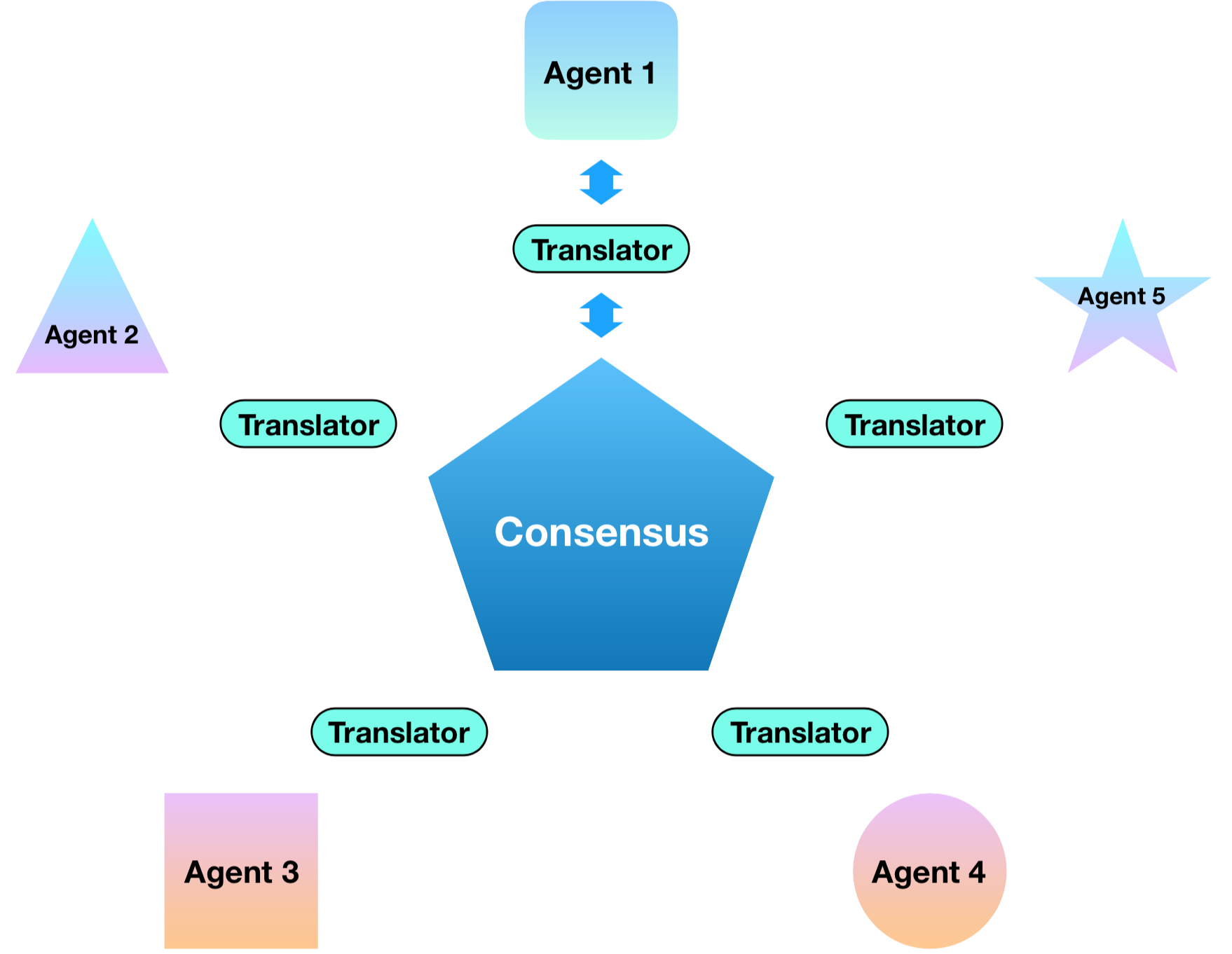}
  \caption{A general framework for heterogeneous federated learning. Each agent owns a private dataset and a uniquely designed model. To communicate and collaborate without data leakage, the agents need to translate their learned knowledge to a standard format. A central server collects these knowledges, compute a consensus distributed across the network. In this work, the translator is implemented using knowledge distillation.}
\end{figure}

\subsection{The framework for heterogeneous federated learning}

We propose FedMD, Algorithm~\ref{algo_HFL}, that solves the problem stated in sec~\ref{sec:ProblemDefinition}. We comment on key components of this framework.
\begin{algorithm}[h]
 \caption{The FedMD framework enabling federated learning for heterogeneous models.}
\KwIn{Public dataset $\mathcal{D}_0$, private datasets $\mathcal{D}_k$,  independently designed model $f_k$, $k=1\dots m$,}
\KwOut {Trained model $f_k$}
\vspace{2pt}
\textbf{Transfer learning:} Each party trains $f_k$ to convergence on the public $\mathcal{D}_0$ and then on its private $\mathcal{D}_k$.\\
\vspace{2pt}
 \For{j=1,2...P} {
  \textbf{Communicate:} Each party computes the class scores $f_k(x_{i}^0)$ on the public dataset, and transmits the result to a central server. \\
   \textbf{Aggregate:} The server computes an updated consensus, which is an average $\tilde{f}(x_{i}^0) = \frac{1}{m}\sum_k f_k(x_{i}^0)$. \\
  \textbf{Distribute:}  Each party downloads the updated consensus $\tilde{f}(x_{i}^0)$. \\
   \textbf{Digest:} Each party trains its model $f_k$ to approach the consensus $\tilde{f}$ on the public dataset $\mathcal{D}_0$.\\
   \textbf{Revisit:} Each party trains its model $f_k$ on its own private data for a few epochs.
   }
\label{algo_HFL}
\end{algorithm}

\vspace*{-0.21in}
\paragraph{Transfer learning:} Before a participant starts the collaboration phase, its model must first undergo the entire transfer learning process. It will be trained fully on the public dataset and then on its own private data. Therefore any future improvements are compared to this baseline. 
\vspace*{-0.1in}
\paragraph{Communication:} We re-purpose the public dataset $\mathcal{D}_0$ as the basis of communication between models, which is realized using knowledge distillation. Each learner $f_k$ expresses its knowledge by sharing the class scores, $f_k(x_{i}^0)$, computed on the public dataset $\mathcal{D}_0$. 
The central server collects these class scores and computes an average $\tilde{f}(x_i^{0})$. 
Each party then trains $f_k$ to approach the consensus  $\tilde{f}(x_i^{0})$.
In this way, the knowledge of one participant can be understood by others without explicitly sharing its private data or model architecture. 
Using the entire large public dataset can cause a large communication burden. 
In practice, the server may randomly select a much smaller subset $d_j \subset \mathcal{D}_0$ at each round as the basis of communication. 
In this way, the cost is under control and does not scale with the complexity of participating models.

\section{Results}
We test this framework in two different environments. In the first environment, the public data is the MNIST and the private data is a subset of the FEMNIST. We consider the i.i.d. case where each private dataset is drawn randomly from FEMNIST, as well as the non-.i.i.d. case where each participant, while only given letters written by a single writer during training, is asked to classify letters by all writers at test time.

In the second environment, the public dataset is the CIFAR10 and the private dataset is a subset of the CIFAR100, which has 100 subclasses that falls under 20 superclasses, e.g. bear, leopard, lion, tiger and wolf belongs to large carnivores. In the i.i.d. case, the task is for each participant to classify test images into correct subclasses. 
The non-i.i.d. case is more challenging:  during training, each participant has data from one subclass per superclass; at test time, participants need to classify generic test data into the correct superclasses. For example, a participant who has only seen wolfs during training is expected to classify lions correctly as large carnivores. 
Therefore it has to rely on information communicated by other participants.

In each environment, 10 participants design unique convolution networks that can differ by number of channels and number of layers, see Table~\ref{table:models_MNIST},\ref{table:models_CIFAR} for details. First they are trained on the public dataset until convergence, --- these models typically have test accuracy around $99\%$ on MNIST and $76\%$ on CIFAR10. Secondly each participant trains its model on its own small private dataset. After these steps, they go through the collaborative training phase, during which the models acquire strong and fast improvements across the board, and quickly outperform the baseline of transfer learning. We use Adam optimizer~\cite{KinBa17} with an initial learning rate of $0.001$; in each round of collaborative training we randomly select a subset $d_j\subset \mathcal{D}_0$ of size 5000 as the basis for communication. More details are given in the supplementary material. The code will be made publicly available after the workshop.

\begin{figure}[H]
  \centering
   \includegraphics[width=0.8\linewidth]{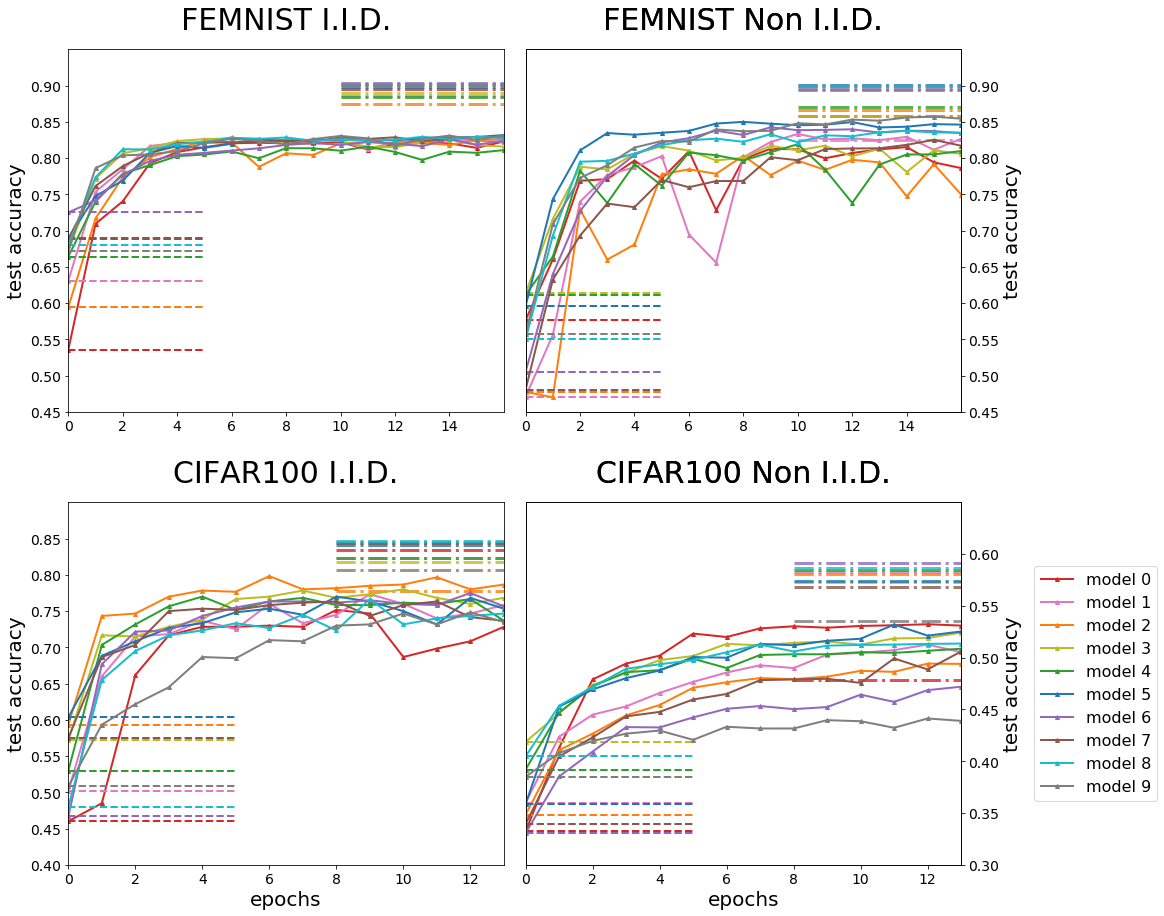}  
  \caption{FedMD improves the test accuracy of participating models beyond their baselines. A dashed line (on the left) represents the test accuracy of a model after full transfer learning with the public dataset and its own small private dataset. This baseline is our starting point and overlaps with the beginning of the corresponding learning curve. A dash-dot line (on the right) represents the would-be performance of a model if private datasets from all participants were declassified and made available to every participant of the group. }
\end{figure}
\vspace*{-0.22in}
\section{Discussion and conclusion}
In this work we proposed FedMD, a framework that enables federated learning for independently designed models. Our framework is based on knowledge distillation and is tested to work on various tasks and datasets. In future we will explore more sophisticated communication module, such as feature transformations and emergent communication protocols that will further improve the performance of our framework. Our framework can also be applied to tasks involving NLP and reinforcement learning. We will extend our framework to extreme cases of heterogeneity involving large discrepancies in the amounts of data, in model capacities and very different local tasks. We believe that heterogeneous federated learning will be an essential tool in future in a broad spectrum of business facing applications of deep learning. 

\section*{Acknowledgments}
We would like to thank Ethan Dyer, Jared Kaplan, Jaehoon Lee, Patrick (Langechuan) Liu, Sam McCandlish, Wenbo Shi, Gennady Voronov, Yunlong Wang, Sho Yaida, Xi Yin and Yao Zhao 
for discussions 
and comments on the manuscript.
DL was supported by the Simons Collaboration Grant on the Non-Perturbative Bootstrap.

\bibliography{HFL}
\bibliographystyle{utphys}

\section*{Supplementary Material}
We provide more details about the models, the datasets, the algorithm and the results in this supplementary material. 

\subsection*{Models}
We list the architectures of the models used by each participant in the MNIST/FEMNIST environment in table~\ref{table:models_MNIST} and those in the CIFAR enrionment in table~\ref{table:models_CIFAR}.

\begin{table}[h]
  \caption{Models for MNIST/FEMNIST}
  \label{sample-table}
  \centering
\begin{tabular}{ |p{1.5cm}||p{2cm}|p{2cm}|p{2cm}|p{1.5cm}|p{2cm}| }
 \hline
Model & 1st conv layer filters ($n_1$)& 2nd conv layer filters ($n_2)$ & 3 conv layer filters ($n_3$) & dropout rate&pre-trained test accuracies on MNIST\\ 
 \hline\hline
0 & 128 & 256 &  None & 0.2&98.6\%\\
1 & 128 & 384 &  None & 0.2&98.8\%\\
2 & 128 & 512 &  None & 0.2&98.4\%\\
3 & 256 & 256 &  None & 0.3 &98.3\%\\
4 & 256 & 512 &  None & 0.4&98.2 \%\\
5 & 64 & 128 &  256 & 0.2&98.9\%\\
6 & 64 & 128 &  192 & 0.2 &99.0\%\\
7 & 128 & 192 &  256 &0.2 &99.1\%\\
8 & 128 & 128 &  128 &0.3&99.2\%\\
9 & 128 & 128 &  192 &0.3&98.9\%\\
 \hline
\end{tabular}
\label{table:models_MNIST}
\end{table}

\begin{table}[h]
  \caption{Models for CIFAR10/CIFAR100}
  \label{sample-table}
  \centering
\begin{tabular}{ |p{1cm}||p{2cm}|p{2cm}|p{2cm}|p{2cm}|p{1.3cm}|p{2cm}|}
 \hline
Model & 1st conv layer filters ($n_1$)& 2nd conv layer filters ($n_2$) & 3rd conv layer filters ($n_3$) 
& 4th conv layer filters ($n_4$) & dropout rate & pre-trained test accuracies on CIFAR10\\ 
 \hline\hline
0 & 128 & 256 &  None & None& 0.2&  71.5\%\\
1 & 128 & 128 &  192 & None & 0.2&  78.8\%\\
2 & 64 & 64 &  64 & None & 0.2&  75.5\%\\
3 & 128 & 64 &  64 & None & 0.3&  74.5\%\\
4 & 64 & 64 & 128 & None & 0.4& 74.9\%\\
5 & 64 & 128 &  256 & None &  0.2 &  74.8\%\\
6 & 64 & 128 &  192 & None & 0.2 &  77.5\%\\
7 & 128 & 192 &  256 & None &0.2 & 77.7\%\\
8 & 128 & 128 &  128 & None & 0.3& 78.8\%\\
9 & 64 & 64 & 64 & 64 & 0.2& 75.4\%\\
 \hline
\end{tabular}
\label{table:models_CIFAR}
\end{table}

\subsection*{Data}
We provide a summary of our public and private datasets in table~\ref{table:data_summary}.

\begin{table}
  \caption{Summary of datasets}
  \label{sample-table}
  \centering
\begin{tabular}{ |p{3cm}||p{2cm}|p{4cm}|p{3cm}| }
 \hline
 Collaborative Task & Public Dataset& Private Classes & Number of Private Data Samples per Class per Party\\ 
 \hline\hline
FEMNIST/MNIST I.I.D. &MNIST  & letters [a-f] classes &  3 \\
FEMNIST/MNIST Non I.I.D.& MNIST & letters from one writers&  around 20 (varies) \\
CIFAR I.I.D.   & CIFAR10 & CIFAR100 subclasses [0,2,20,63,71,82] &  3\\
CIFAR Non I.I.D.  &  CIFAR10 & CIFAR100 superclasses [0-5]  & 20\\
 \hline
\end{tabular}
\label{table:data_summary}
\end{table}

\subsection*{Method}
We clarify important details about our implementation of Algorithm \ref{algo_HFL}:
\begin{enumerate}
\item In the communication phase, the models communicate and align their logits computed from public data without applying the softmax activation layer . We could also use the softmax score with a particular temperature \cite{Hinton44873}, and we do not expect large effects from this distinction. 
\item In the communication phase, instead of using the entire public dataset, we use a subset of size 5000 that is randomly selected at each round. This speeds up the process without sacrificing the performance.
\item The number of rounds and the batch size in the Digest and the Revisit phase control the stability of the learning process. A model may undergo transient retrogression in test performance that is quickly recovered in the next couple of rounds. This issue can be resolved by choosing smaller number of epochs in the revisit phase and larger batch size in the digest phase. 
\item In principle the consensus can be computed using a weighted average $\tilde{f}(x_{i}^0) = \sum_k c_k f_k(x_{i}^0)$. In this work we almost always choose the weights $c_k$ to be equal to $1/N_{parties}$. One exception is in the CIFAR case where we slightly suppress the contribution from two weaker models (0 and 9). These weights may become more important when we have extremely different models or data. 
\end{enumerate}

\subsection*{Results}
We discuss several interesting aspects of our results.
\begin{enumerate}
\item We measure our results against the test accuracy that a model could have achieved if the private data of all participants were pooled and made directly available to the whole group. See Table~\ref{table:ub}. Usually our framework boosts the performance of all participants to a level only a few percent lower than this pooled data performance.

\item There are isolated cases where a model trained in our framework consistently outperforms the same model trained with pooled private data. In particular model-0 in the CIFAR non-i.i.d. case. Besides, its performance is mostly on the top of the herd. This model has the simplest architecture and is usually lagging behind its more sophisticated peers. It is interesting to understand the mechanism behind this success and utilize it to improve our framework. 

\item Our framework can incorporate extreme cases of model heterogeneity. We have experimented with several models having much lower performance, such as two layer fully connected networks. If they contribute to the consensus with the same weight as the advanced models, they tend to hinder the accuracy of the herd. Our framework works better if we suppress their contribution with a lower weight. 
\end{enumerate}

\begin{table}[h]
  \caption{Performance of models trained with pooled private data.}
  \label{sample-table}
  \centering
\begin{tabular}{ |p{3cm}||p{9.8cm}| }
 \hline
 Collaborative Task & Each model's performance trained with pooled private data\\ 
 \hline\hline
FEMNIST/MNIST I.I.D. &  $[0. 895, 
0. 886, 
0. 875,
0. 889,
0. 885,
0. 899
0. 903,
0. 902,
0. 902,
0.901]$\\
FEMNIST/MNIST Non I.I.D.& $[0.858, 0.825, 0.867, 0.858, 0.870, 0.901, 0.896, 0.899, 0.900, 0.894]$\\
CIFAR I.I.D.  & $[0.835, 0.823, 0.778, 0.818, 0.823, 0.842, 0.845, 0.843, 0.847,
       0.807]$ \\
CIFAR Non I.I.D.  & $ [0.478, 0.583, 0.581, 0.573, 0.584, 0.574, 0.591, 0.568, 0.586,
       0.535]$\\
 \hline
\end{tabular}
\label{table:ub}
\end{table}

\end{document}